\pdfoutput=1

\documentclass[11pt]{article}

\usepackage{ACL2023}
\usepackage{times}
\usepackage{latexsym}

\usepackage[T1]{fontenc}

\usepackage[utf8]{inputenc}

\usepackage{microtype}

\usepackage{graphicx}
\usepackage{multicol}
\usepackage{multirow}
\usepackage{amsmath}
\usepackage{amssymb}
\usepackage{bbding}
\usepackage{subfigure}
\usepackage{colortbl}
\usepackage{pifont}
\usepackage{bbm}
\usepackage{makecell}
\usepackage{bbding}
\usepackage{mathrsfs}
\usepackage{bm}
\usepackage{CJKutf8}
\usepackage{booktabs}
\usepackage{tabu}
\usepackage{fontawesome}
\usepackage{mdframed}
\usepackage[linesnumbered,ruled,vlined]{algorithm2e}
\usepackage{tcolorbox}
\NewDocumentCommand{\zhangwei}
{ mO{} }{\textcolor{blue}{\textsuperscript{\textit{zhangwei}}\textsf{\textbf{\small[#1]}}}}

\newcommand{\sftmodel}{UCoder}
\newcommand{\ourmethod}{IPC} 

\usepackage{amsthm}
\theoremstyle{plain} 
\theoremstyle{definition} 
\usepackage{amsthm}
\usepackage{enumerate}
\newtheorem{theorem}{Theorem}[section]
\newtheorem{definition}[theorem]{Definition}

\title{\sftmodel{}: Unsupervised Code Generation by Internal \\ Probing of Large Language Models}

\author{
  Jiajun Wu\textsuperscript{\rm 1},
  {\bf Jian Yang}\textsuperscript{\rm 1\thanks{\ \ Corresponding Author.}},
  {\bf Wei Zhang}\textsuperscript{\rm 1},
  {\bf Lin Jing}\textsuperscript{\rm 1},
  {\bf Yuqing Ma}\textsuperscript{\rm 2},
  {\bf Ensheng Shi}\textsuperscript{\rm 2},  \\
  {\bf Yuchi Ma}\textsuperscript{\rm 2},
  {\bf Zhoujun Li}\textsuperscript{\rm 2},
  {\bf Xianglong Liu}\textsuperscript{\rm 1} \\
   \textsuperscript{\rm 1}Beihang University;
   \textsuperscript{\rm 2}Huawei; \\
   \texttt{\{wuyuverse,jiayang\}@buaa.edu.cn} \\
}
\newmdenv[
  backgroundcolor=red!05,
  linecolor=quoteborder,
  skipabove=1em,
  skipbelow=0em,
  leftline=true,
  topline=false,
  bottomline=false,
  rightline=false,
  linecolor=red!66,
  linewidth=4pt
]{githubquote}
\begin{document}
\begin{CJK*}{UTF8}{gbsn}
\maketitle

\begin{abstract}
Large language models (LLMs) have demonstrated remarkable capabilities in code generation tasks. However, their effectiveness heavily relies on supervised training with extensive labeled (e.g., question-answering pairs) or unlabeled datasets (e.g., code snippets), which are often expensive and difficult to obtain at scale. To address this limitation, this paper introduces a method \textbf{\ourmethod{}}, an unsupervised framework that leverages \textbf{I}nternal \textbf{P}robing of LLMs for \textbf{C}ode generation without any external corpus, even unlabeled code snippets. We introduce the problem space probing, test understanding probing, solution space probing, and knowledge consolidation and reinforcement to probe the internal knowledge and confidence patterns existing in LLMs. Further, \ourmethod{} identifies reliable code candidates through self-consistency mechanisms and representation-based quality estimation to train \sftmodel{} (coder with unsupervised learning). We validate the proposed approach across multiple code benchmarks, demonstrating that unsupervised methods can achieve competitive performance compared to supervised approaches while significantly reducing the dependency on labeled data and computational resources. Analytic experiments reveal that internal model states contain rich signals about code quality and correctness, and that properly harnessing these signals enables effective unsupervised learning for code generation tasks, opening new directions for training code LLMs in resource-constrained scenarios.
\end{abstract}

\section{Introduction}
Large language models (LLMs) have demonstrated strong capabilities in code generation, producing functional code from natural language descriptions. This progress has attracted substantial interest from both academia and industry due to its practical impact on software development. Closed-source LLMS such as GPT-5~\cite{openai2025gpt5codex} and Claude-4.5~\cite{claude45} can generate file-level code with high accuracy, while open-source alternatives, including StarCoder~\cite{starcoder,starcoder2}, DeepSeek-Coder~\cite{deepseek_coder}, and QwenCoder~\cite{qwen25coder} have emerged as competitive solutions for code intelligence.

\begin{figure}[t]
\centering
\includegraphics[width=0.9\columnwidth]{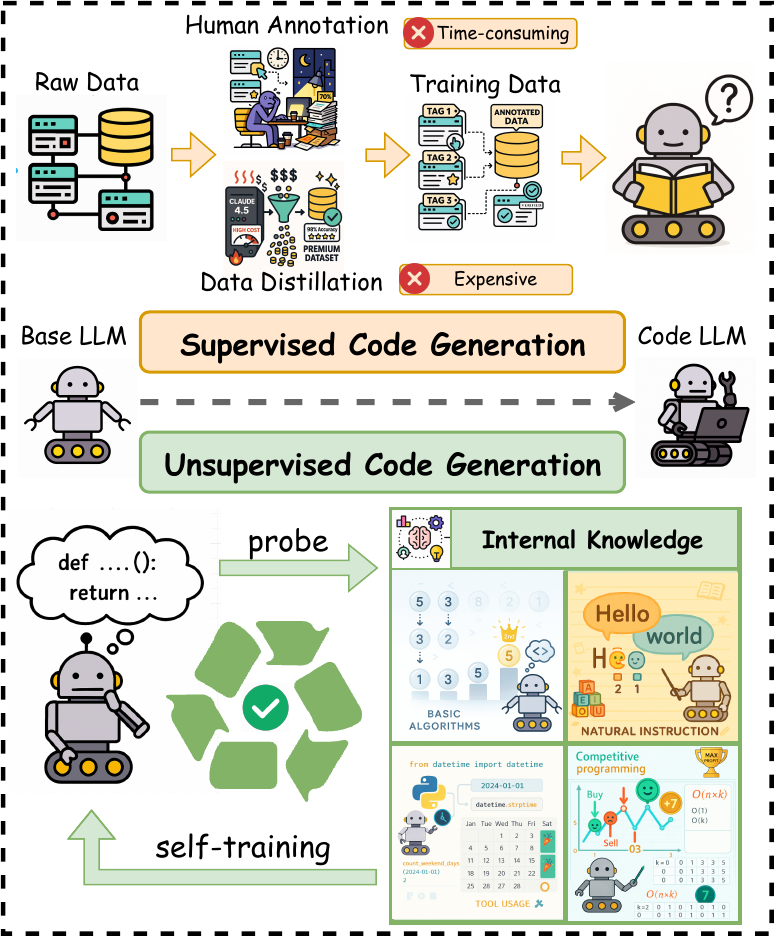}
\caption{Comparison between supervised and unsupervised paradigms for code generation.}
\label{fig:paradigm_comparison}
\vspace{-20pt}
\end{figure}
\begin{figure*}[t]
\centering
\includegraphics[width=0.95\textwidth]{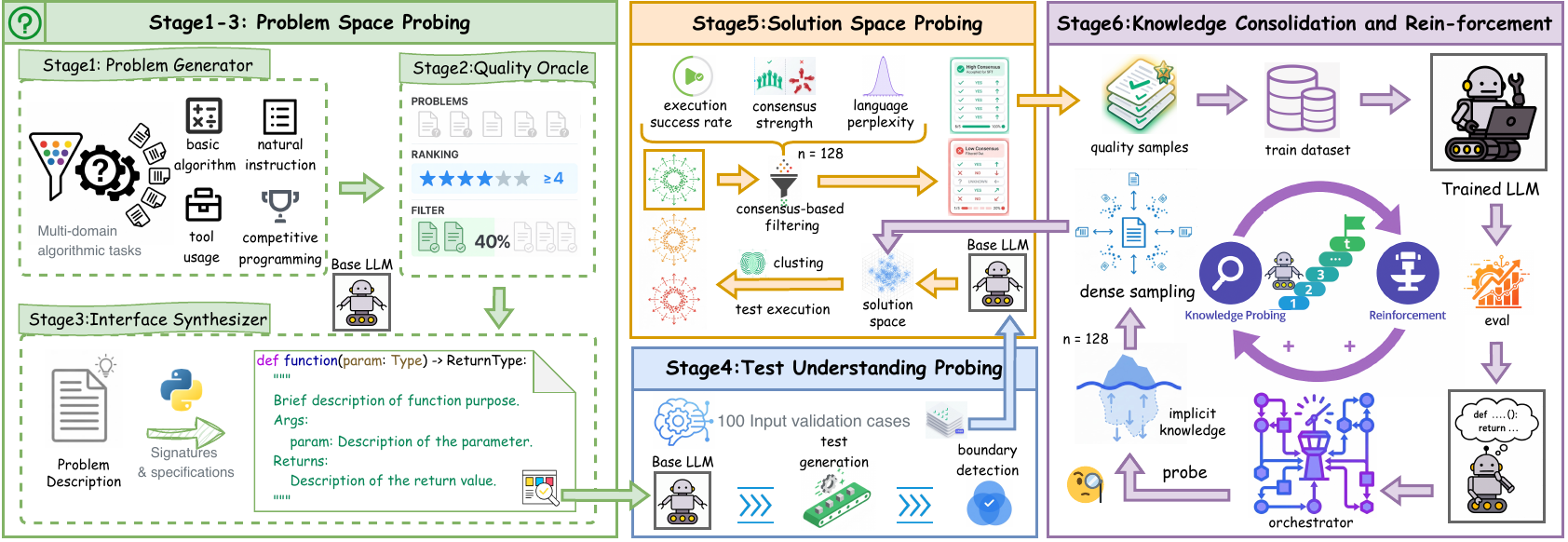}
\caption{Overview of the proposed six-stage self-bootstrapping framework for unsupervised code generation.}
\label{fig:framework_overview}
\vspace{-15pt}
\end{figure*}

Most existing approaches for improving code generation rely on supervised instruction tuning, where LLMs are fine-tuned on curated problem-solution pairs annotated by human experts or LLM-based annotated. However, creating high-quality instruction data requires substantial human effort in problem design, implementation, and verification, with costs increasing as model capabilities advance. 
The recent~\cite{sft_effect1,sft_effect2,sft_effect3} works emphasize that pre-training brings the knowledge, and post-training is weak at knowledge integration but focuses on knowledge utilization and alignment. \textit{These challenges motivate a fundamental question: Can LLMs autonomously improve their generation capabilities using post-training without any external corpus, relying only on pre-trained knowledge?}

In this work, we introduce an unsupervised framework that performs \textbf{I}nternal \textbf{P}robing of LLMs for \textbf{C}ode generation, enabling post-training without external corpora or human-annotated instruction data. Our approach exploits latent programming knowledge in LLMs and uses execution feedback as a scalable, deterministic supervision signal grounded in program semantics. We implement a six-stage self-bootstrapping process that generates diverse programming tasks, synthesizes test suites, samples candidate solutions, and applies execution-driven consensus clustering to identify correct implementations. High-consensus solutions are iteratively consolidated as training data, forming a feedback loop that progressively improves model performance.

Despite using no external data, \sftmodel{} achieves comparable performance to the supervised baseline across multiple benchmarks. The primary contributions of this work are:
\begin{itemize}
\setlength\itemsep{0em}
\item We successfully probe latent programming knowledge in LLMs by forcing models to generate programming problems and their solutions, then identify correct solutions by finding clusters of similar implementations. Then, the self-training method progressively improves the LLM by reinforcing solutions. 
\item Based on the self-generated data from the unsupervised framework using internal probing of LLMs (\ourmethod{}) without any external data, \sftmodel{} (7B, 14B, 32B) achieves performance competitive with supervised baselines.
\item We provide empirical analysis showing that self-generated data maintains rich lexical, semantic, and structural diversity, while consensus-based selection improves solution quality and exhibits inverse scaling behavior.
\end{itemize}

\section{Unsupervised Code Generation}
\subsection{Task Definition}

\subsubsection{Supervised Code Generation}
Supervised code generation is formulated as a sequence-to-sequence learning task. Given a training dataset $\mathcal{D} = \{(x_i, y_i^)\}_{i=1}^N$, the model parameters are optimized by maximizing log-likelihood:\begin{MiddleEquation}\begin{equation}\theta^* = \arg\max_\theta \sum_{i=1}^N \log p_\theta(y_i|x_i)\end{equation}\end{MiddleEquation}where $x_i \in \mathcal{X}$ denotes a natural language query, $y_i \in \mathcal{Y}$ represents the corresponding reference implementation, and $p_\theta(y|x)$ is the conditional distribution over code sequences parameterized by $\theta$. We adopt Pass@k~\cite{chen2021codex} to evaluate code correctness, which measures the probability that at least one of $k$ independently sampled solutions passes all test cases.

\subsubsection{Unsupervised Code Generation}

Unsupervised code generation aims to improve LLM code generation capabilities without human-annotated supervision. Given an initial model $M_0: \mathcal{X} \rightarrow \mathcal{Y}$, the objective is to develop a self-improvement algorithm $\mathcal{A}$ producing an enhanced model $M^* = \mathcal{A}(M_0)$ such that $\texttt{Pass@}k(M^*) > \texttt{Pass@}k(M_0)$ on held-out test sets, without access to paired training data $(x, y) \in \mathcal{X} \times \mathcal{Y}$. This presents three fundamental challenges: (1) \textbf{Problem Space Construction.} Automatically generating diverse programming problems with appropriate difficulty distributions while maintaining semantic clarity; (2) \textbf{Unsupervised Correctness Verification.} Assessing functional correctness without reference implementations; (3) \textbf{Self-Bootstrapping Signal Construction.} Extracting reliable training signals from noisy candidates while ensuring iterative stability. We address these through an execution-driven consensus mechanism coupled with a self-bootstrapping framework, detailed in the following sections.

\subsection{Probing Internal Knowledge in LLMs}
\begin{figure}[t]
\centering
\includegraphics[width=1.0\columnwidth]{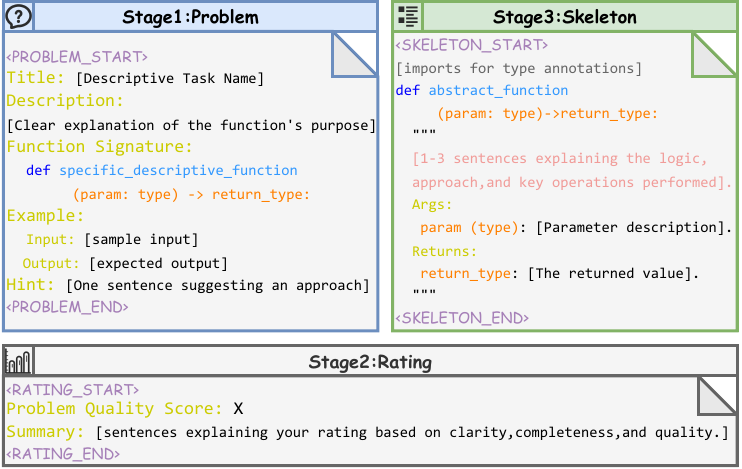}
\caption{Problem space probing proceeds through three stages: problem generation with function signatures and input-output contracts, difficulty rating assessment and categorization, and solution skeleton generation with implementation structure.}
\label{fig:combined_examples}
\vspace{-10pt}
\end{figure}

LLMs encode extensive programming knowledge through pre-training, yet this knowledge remains implicit and difficult to elicit. We propose a six-stage framework to surface and reinforce these latent capabilities, as shown in \autoref{fig:framework_overview}. First, we probe the problem space (Stages~1--3) by prompting the model to generate algorithmic problems with complete specifications, revealing its understanding of programming paradigms and data structures. Representative examples of problem generation, difficulty assessment, and solution skeleton construction are illustrated in \autoref{fig:combined_examples}. 

We then assess semantic understanding (Stage~4) by generating approximately 100 test cases per problem to identify boundary conditions and edge cases. At the core (Stage~5), we probe the solution space via dense sampling, where execution-driven consensus clustering reveals that correct implementations form tight clusters while incorrect ones are dispersed. We quantify solution quality using execution success rate $e(r)$, consensus strength $s(r)$, and code fluency $f(r)$. Finally (Stage~6), we consolidate high-consensus samples through supervised fine-tuning, reinforcing correct patterns. The process forms a positive feedback loop: at iteration $t$, the improved LLM $M_t$ produces higher-quality candidates, enabling more reliable selection and further strengthening $M_{t+1}$. Our experimental results in \autoref{subsec:results} validate that pre-trained models already contain the knowledge required to solve target tasks in implicit form.

\subsection{Execution-Driven Consensus Clustering}


Our approach exploits that correctness is singular while incorrectness is diverse: correct implementations produce identical outputs, but incorrect ones fail heterogeneously. This clustering structure allows the maximum-consensus cluster to indicate correctness without ground truth, formalized in \autoref{thm:consensus} and validated in \autoref{sec:consensus_analysis}.

\subsubsection{Definitions}
Consensus clustering has three definitions.

\definecolor{HeaderColor}{HTML}{9CA3AF}        
\definecolor{Size7BColor}{HTML}{F3E5F5}        
\definecolor{Size14BColor}{HTML}{E8F5E8}       
\definecolor{Size32BColor}{HTML}{F9FBE7}       
\definecolor{OursColor}{HTML}{E3F2FD}          
\definecolor{ClosedAPIColor}{HTML}{F5F5F5}     


\begin{table*}[t]
  \centering
  \resizebox{0.95\textwidth}{!}{%
    \begin{tabular}{l|cc|cc|cc|cc|c|c}
      \toprule
      \multirow{2}{*}{\textbf{Model}} & 
      \multicolumn{2}{c|}{\textbf{HumanEval}} & 
      \multicolumn{2}{c|}{\textbf{MBPP}} & 
      \multicolumn{2}{c|}{{\textbf{BCB-Complete}}} & 
      \multicolumn{2}{c|}{{\textbf{BCB-Instruct}}} & 
      \textbf{LiveCode-} & 
      \textbf{FullStack-} \\
      & HE & HE+ & MBPP & MBPP+ & Full & Hard & Full & Hard & \textbf{Bench} & \textbf{Bench} \\
      \rowcolor{HeaderColor}\multicolumn{11}{c}{\rule{0pt}{2.5ex}\textcolor{white}{\textbf{6B+ LLMs}}\rule[-1ex]{0pt}{1ex}} \\
      \rowcolor{Size7BColor} \small CodeLlama-7B-Instruct & 40.9 & 33.5 & 39.9 & 33.6 & 25.7 & 4.1 & 21.9 & 3.4 & 7.1 & 25.40 \\
      \rowcolor{Size7BColor} \small DS-Coder-6.7B-Instruct & 74.4 & 71.3 & 74.9 & 65.6 & 43.8 & 15.5 & 35.5 & 10.1 & 15.5 & 40.16 \\
      \rowcolor{Size7BColor} \small OpenCoder-8B-Instruct & 83.5 & 78.7 & 79.1 & 69.0 & 50.9 & 18.9 & \textbf{43.2} & \textbf{18.2} & \textbf{23.2} & 41.08 \\
      \rowcolor{Size7BColor} \small Qwen2.5-Coder-7B          & 61.6 & 53.0 & 76.9 & 62.9 & 45.8 & 16.2 & - & - & - & - \\
      \rowcolor{Size7BColor} \small Qwen2.5-Coder-7B-Instruct & \textbf{88.4} & \textbf{84.1} & 83.5 & 71.7 & 48.8 & 20.3 & 41.0 & \textbf{18.2} & 18.2 & 47.95 \\
      \rowcolor{OursColor} \small \textbf{\sftmodel{}-7B} & 83.5 & 76.8 & \textbf{85.2} & \textbf{72.2} & \textbf{52.0} & \textbf{22.3} & 41.1 & 15.5 & 22.9 & \textbf{51.27} \\
      \rowcolor{HeaderColor}\multicolumn{11}{c}{\rule{0pt}{2.5ex}\textcolor{white}{\textbf{13B+ LLMs}}\rule[-1ex]{0pt}{1ex}} \\
      \rowcolor{Size14BColor} \small CodeLlama-13B-Instruct & 40.2 & 32.3 & 60.3 & 51.1 & 31.7 & 6.8 & 28.5 & 9.5 & 6.1 & 27.00 \\
      \rowcolor{Size14BColor} \small Starcoder2-15B-Instruct-v0.1 & 67.7 & 60.4 & 78.0 & 65.1 & 45.1 & 14.9 & 37.2 & 11.5 & 12.1 & 42.68 \\
      \rowcolor{Size14BColor} \small DS-Coder-V2-Lite-Instruct & 81.1 & 75.6 & 82.8 & 70.4 & 47.6 & 18.2 & 36.8 & 16.2 & \textbf{24.3} & - \\
      \rowcolor{Size14BColor} \small Qwen2.5-Coder-14B          & 64.0 & 57.9 & 81.0 & 66.7 & 51.8 & 22.3 & - & - & - & - \\
      \rowcolor{Size14BColor} \small Qwen2.5-Coder-14B-Instruct & \textbf{89.6} & \textbf{87.2} & 86.2 & 72.8 & \textbf{56.7} & \textbf{29.7} & \textbf{48.4} & \textbf{22.2} & 23.4 & \textbf{55.28} \\
      \rowcolor{OursColor} \small \textbf{\sftmodel{}-14B} & 87.8 & 81.1 & \textbf{86.5} & \textbf{74.3} & 53.9 & 24.3 & 40.9 & 16.2 & 20.6 & 52.52 \\
      \rowcolor{HeaderColor}\multicolumn{11}{c}{\rule{0pt}{2.5ex}\textcolor{white}{\textbf{32B+ LLMs}}\rule[-1ex]{0pt}{1ex}} \\
      \rowcolor{Size32BColor} \small CodeLlama-34B-Instruct & 48.2 & 40.2 & 61.1 & 50.5 & 35.6 & 10.8 & 29.0 & 8.8 & 8.4 & 27.56 \\
      \rowcolor{Size32BColor} \small DS-Coder-33B-Instruct & 81.1 & 75.0 & 80.4 & 70.1 & 51.1 & 20.9 & 42.0 & 17.6 & 21.3 & 48.19 \\
      \rowcolor{Size32BColor} \small DS-Coder-V2-Instruct & 85.4 & 82.3 & 89.4 & 75.1 & \textbf{59.7} & 29.7 & 48.2 & 24.3 & 27.9 & 56.37 \\
      \rowcolor{Size32BColor} \small Qwen2.5-Coder-32B          & 65.9 & 60.4 & 83.0 & 68.2 & 53.6 & 26.4 & - & - & - & - \\
      \rowcolor{Size32BColor} \small Qwen2.5-Coder-32B-Instruct & \textbf{92.7} & \textbf{87.2} & \textbf{90.2} & 75.1 & 58.0 & \textbf{33.8} & \textbf{49.6} & \textbf{27.0} & \textbf{31.4} & \textbf{56.88} \\
      \rowcolor{OursColor} \small \textbf{\sftmodel{}-32B} & 89.0 & 82.9 & 89.7 & \textbf{75.7} & 55.4 & 27.7 & 45.7 & 17.6 & 21.4 & 53.35 \\
      \rowcolor{HeaderColor}\multicolumn{11}{c}{\rule{0pt}{2.5ex}\textcolor{white}{\textbf{Closed-APIs}}\rule[-1ex]{0pt}{1ex}} \\
      \rowcolor{ClosedAPIColor} \small GPT-4o-2024-08-06 & 92.1 & 86.0 & 86.8 & 72.5 & - & 36.5 & 50.1 & 25.0 & 34.6 & 58.89 \\
      \rowcolor{ClosedAPIColor} \small Claude-3.5-Sonnet-20241022 & 92.1 & 86.0 & 91.0 & 74.6 & 58.6 & 35.1 & 46.8 & 25.7 & 31.6 & 60.70 \\
      \midrule
    \end{tabular}%
  }%
  \caption{Performance comparison of \textbf{Qwen2.5-Coder} Base and Instruct models with our iterative SFT models across code generation benchmarks. All metrics represent Pass@1 execution rates (\%). Complete split is reported for Base models and Instruct split for Instruct models. \textbf{Bold} indicates best performance within each size category. ``-'' denotes unavailable or inapplicable results.}
  \label{tab:qwen25_base_instruct_comparison}
  \vspace{-15pt}
\end{table*}

\begin{definition}[Execution Signature]
Given candidates $R=\{r_1,\dots,r_n\}$ and tests $T=\{t_1,\dots,t_m\}$, define
$\mathrm{Exec}: R\times T \rightarrow \{0,1\}$ as the pass indicator:
$\mathrm{Exec}(r_i,t_j)=1$ if $r_i$ passes $t_j$, and $0$ otherwise.
The execution signature of $r_i$ on $T$ is
\begin{MiddleEquation}
\begin{equation}
\sigma(r_i;T)=\bigoplus_{j=1}^{m}\mathrm{Exec}(r_i,t_j),
\end{equation}
\end{MiddleEquation}
where $\bigoplus$ denotes ordered concatenation. Thus $\sigma(r_i;T)\in\{0,1\}^{m}$, and
$\sigma(r_i;T)=\mathbf{1}_m$ indicates that $r_i$ passes all $m$ tests.
\end{definition}

\begin{definition}[Consensus Clusters]
If $\sigma(r_i,T)=\sigma(r_j,T)$, we can regard the $r_{i}$ and $r_{j}$ as equivalent solution.
Given the value of $\sigma(r_i,T)$, we can partition $R$ into clusters $\mathcal{C}=\{C_1,\dots,C_\ell\}$ of behaviorally identical candidates.
\end{definition}

\begin{definition}[Quality Metrics]
Each candidate $r\in R$ is scored by:
\begin{equation}
\begin{aligned}
e(r) &= \frac{|\{t\in T:\mathrm{Exec}(r,t)\neq\perp\}|}{|T|}, \\
s(r) &= |\{r'\in R:\sigma(r')=\sigma(r)\}|, \\
f(r) &= \exp\!\left(-\frac{1}{|r|}\sum_{i=1}^{|r|}\log p(x_i\mid x_{<i})\right),
\end{aligned}
\end{equation}where $e(r)$ measures execution success, $s(r)$ consensus strength, and $f(r)$ code fluency.
\end{definition}

\subsubsection{Hierarchical Selection}
\definecolor{HeaderColor}{HTML}{9CA3AF}        
\definecolor{Size7BColor}{HTML}{F3E5F5}        
\definecolor{Size14BColor}{HTML}{E8F5E8}       
\definecolor{Size32BColor}{HTML}{F9FBE7}       
\definecolor{BestIterColor}{HTML}{E3F2FD}      
\definecolor{BaselineColor}{HTML}{FFE0B2}      

\newcolumntype{C}{w{c}{1.1cm}}

\begin{table*}[t]
  \centering
  \scriptsize
  \renewcommand{\arraystretch}{1.15}
  \setlength{\tabcolsep}{2pt}
  \resizebox{0.95\textwidth}{!}{%
    \begin{tabular}{c|CC|CC|CC|CC|c|c}
    \toprule
      \multirow{2}{*}{\textbf{Iter}} & 
      \multicolumn{2}{c|}{\textbf{HumanEval}} & 
      \multicolumn{2}{c|}{\textbf{MBPP}} & 
      \multicolumn{2}{c|}{{\textbf{BCB-Complete}}} & 
      \multicolumn{2}{c|}{{\textbf{BCB-Instruct}}} & 
      {\textbf{LiveCode-}} & 
      {\textbf{FullStack-}} \\
      & HE & HE+ & MBPP & MBPP+ & Full & Hard & Full & Hard & {\textbf{Bench}} & {\textbf{Bench}} \\
      \rowcolor{HeaderColor}\multicolumn{11}{c}{\rule{0pt}{2.5ex}\textcolor{white}{\textbf{Ucoder-7B}}\rule[-0.6ex]{0pt}{0pt}} \\
      \rowcolor{BaselineColor} 0 & 77.4& 67.1& 72.0& 63.0& 44.4& 15.5& 34.6& 14.9& 13.0& 40.2 \\
      \rowcolor{Size7BColor} 1 & 81.7\tiny$_{\color{green!60!black}\text{+}4.3}$& 74.4\tiny$_{\color{green!60!black}\text{+}7.3}$& 72.0\tiny$_{\color{gray}0.0}$& 63.0\tiny$_{\color{gray}0.0}$& 51.3\tiny$_{\color{green!60!black}\text{+}6.9}$& \textbf{23.0}\tiny$_{\color{green!60!black}\text{+}7.5}$& \textbf{42.2}\tiny$_{\color{green!60!black}\text{+}7.6}$& \textbf{20.9}\tiny$_{\color{green!60!black}\text{+}6.0}$& 15.3\tiny$_{\color{green!60!black}\text{+}2.3}$& 48.2\tiny$_{\color{green!60!black}\text{+}7.9}$ \\
      \rowcolor{Size7BColor} 2 & \textbf{84.1}\tiny$_{\color{green!60!black}\text{+}6.7}$& \textbf{77.4}\tiny$_{\color{green!60!black}\text{+}10.3}$& 79.1\tiny$_{\color{green!60!black}\text{+}7.1}$& 66.4\tiny$_{\color{green!60!black}\text{+}3.4}$& 44.7\tiny$_{\color{green!60!black}\text{+}0.3}$& 14.9\tiny$_{\color{red}\text{-}0.6}$& 35.8\tiny$_{\color{green!60!black}\text{+}1.2}$& 12.2\tiny$_{\color{red}\text{-}2.7}$& 14.5\tiny$_{\color{green!60!black}\text{+}1.5}$& 40.2\tiny$_{\color{red}\text{-}0.1}$ \\
      \rowcolor{Size7BColor} 3 & \textbf{84.1}\tiny$_{\color{green!60!black}\text{+}6.7}$& \underline{76.8}\tiny$_{\color{green!60!black}\text{+}9.7}$& 81.2\tiny$_{\color{green!60!black}\text{+}9.2}$& 69.0\tiny$_{\color{green!60!black}\text{+}6.0}$& 44.4\tiny$_{\color{gray}0.0}$& 15.5\tiny$_{\color{gray}0.0}$& 34.5\tiny$_{\color{red}\text{-}0.1}$& 13.5\tiny$_{\color{red}\text{-}1.4}$& \underline{21.4}\tiny$_{\color{green!60!black}\text{+}8.4}$& 40.2\tiny$_{\color{gray}0.0}$ \\
      \rowcolor{Size7BColor} 4 & \textbf{84.1}\tiny$_{\color{green!60!black}\text{+}6.7}$& \textbf{77.4}\tiny$_{\color{green!60!black}\text{+}10.3}$& 79.1\tiny$_{\color{green!60!black}\text{+}7.1}$& 66.4\tiny$_{\color{green!60!black}\text{+}3.4}$& 44.7\tiny$_{\color{green!60!black}\text{+}0.3}$& 14.9\tiny$_{\color{red}\text{-}0.6}$& 35.8\tiny$_{\color{green!60!black}\text{+}1.2}$& 12.2\tiny$_{\color{red}\text{-}2.7}$& 14.5\tiny$_{\color{green!60!black}\text{+}1.5}$& 40.2\tiny$_{\color{red}\text{-}0.1}$ \\
      \rowcolor{Size7BColor} 5 & 81.7\tiny$_{\color{green!60!black}\text{+}4.3}$& 75.0\tiny$_{\color{green!60!black}\text{+}7.9}$& \underline{83.9}\tiny$_{\color{green!60!black}\text{+}11.9}$& \underline{71.2}\tiny$_{\color{green!60!black}\text{+}8.2}$& \textbf{52.2}\tiny$_{\color{green!60!black}\text{+}7.8}$& 19.6\tiny$_{\color{green!60!black}\text{+}4.1}$& 40.7\tiny$_{\color{green!60!black}\text{+}6.1}$& 14.2\tiny$_{\color{red}\text{-}0.7}$& 20.6\tiny$_{\color{green!60!black}\text{+}7.6}$& \underline{50.0}\tiny$_{\color{green!60!black}\text{+}9.7}$ \\
      \rowcolor{BestIterColor} \textbf{6} & \underline{83.5}\tiny$_{\color{green!60!black}\text{+}6.1}$& \underline{76.8}\tiny$_{\color{green!60!black}\text{+}9.7}$& \textbf{85.2}\tiny$_{\color{green!60!black}\text{+}13.2}$& \textbf{72.2}\tiny$_{\color{green!60!black}\text{+}9.2}$& \underline{52.0}\tiny$_{\color{green!60!black}\text{+}7.6}$& \underline{22.3}\tiny$_{\color{green!60!black}\text{+}6.8}$& \underline{41.1}\tiny$_{\color{green!60!black}\text{+}6.5}$& \underline{15.5}\tiny$_{\color{green!60!black}\text{+}0.6}$& \textbf{22.9}\tiny$_{\color{green!60!black}\text{+}9.9}$& \textbf{51.3}\tiny$_{\color{green!60!black}\text{+}11.0}$ \\
      \rowcolor{HeaderColor}\multicolumn{11}{c}{\rule{0pt}{1.8ex}\textcolor{white}{\textbf{Ucoder-14B}}\rule[-0.6ex]{0pt}{0pt}} \\
      \rowcolor{BaselineColor} 0 & 83.5& 76.8& 75.9& 64.0& \underline{53.3}& \underline{23.6}& \underline{43.2}& \textbf{16.2}& \underline{22.1}& 50.1 \\
      \rowcolor{Size14BColor} 1 & 85.4\tiny$_{\color{green!60!black}\text{+}1.9}$& 77.4\tiny$_{\color{green!60!black}\text{+}0.6}$& \textbf{87.8}\tiny$_{\color{green!60!black}\text{+}11.9}$& \underline{73.3}\tiny$_{\color{green!60!black}\text{+}9.3}$& 53.1\tiny$_{\color{red}\text{-}0.2}$& 21.6\tiny$_{\color{red}\text{-}2.0}$& 41.0\tiny$_{\color{red}\text{-}2.2}$& 14.2\tiny$_{\color{red}\text{-}2.0}$& 17.6\tiny$_{\color{red}\text{-}4.5}$& \textbf{53.6}\tiny$_{\color{green!60!black}\text{+}3.5}$ \\
      \rowcolor{Size14BColor} 2 & 84.8\tiny$_{\color{green!60!black}\text{+}1.3}$& 76.2\tiny$_{\color{red}\text{-}0.6}$& 84.9\tiny$_{\color{green!60!black}\text{+}9.0}$& 70.6\tiny$_{\color{green!60!black}\text{+}6.6}$& 50.4\tiny$_{\color{red}\text{-}2.9}$& \underline{23.6}\tiny$_{\color{gray}0.0}$& 40.9\tiny$_{\color{red}\text{-}2.3}$& \underline{15.5}\tiny$_{\color{red}\text{-}0.7}$& 19.1\tiny$_{\color{red}\text{-}3.0}$& 49.9\tiny$_{\color{red}\text{-}0.2}$ \\
      \rowcolor{Size14BColor} 3 & 84.8\tiny$_{\color{green!60!black}\text{+}1.3}$& 78.0\tiny$_{\color{green!60!black}\text{+}1.2}$& 83.6\tiny$_{\color{green!60!black}\text{+}7.7}$& 72.2\tiny$_{\color{green!60!black}\text{+}8.2}$& 51.6\tiny$_{\color{red}\text{-}1.7}$& 23.0\tiny$_{\color{red}\text{-}0.6}$& 41.8\tiny$_{\color{red}\text{-}1.4}$& \underline{15.5}\tiny$_{\color{red}\text{-}0.7}$& 18.3\tiny$_{\color{red}\text{-}3.8}$& 49.8\tiny$_{\color{red}\text{-}0.3}$ \\
      \rowcolor{Size14BColor} 4 & 84.8\tiny$_{\color{green!60!black}\text{+}1.3}$& 78.0\tiny$_{\color{green!60!black}\text{+}1.2}$& 84.1\tiny$_{\color{green!60!black}\text{+}8.2}$& 72.8\tiny$_{\color{green!60!black}\text{+}8.8}$& 51.8\tiny$_{\color{red}\text{-}1.5}$& 18.2\tiny$_{\color{red}\text{-}5.4}$& 41.1\tiny$_{\color{red}\text{-}2.1}$& 12.8\tiny$_{\color{red}\text{-}3.4}$& \underline{22.1}\tiny$_{\color{gray}0.0}$& 50.4\tiny$_{\color{green!60!black}\text{+}0.3}$ \\
      \rowcolor{BestIterColor} \textbf{5} & \textbf{87.8}\tiny$_{\color{green!60!black}\text{+}4.3}$& \textbf{81.1}\tiny$_{\color{green!60!black}\text{+}4.3}$& \underline{86.5}\tiny$_{\color{green!60!black}\text{+}10.6}$& \textbf{74.3}\tiny$_{\color{green!60!black}\text{+}10.3}$& \textbf{53.9}\tiny$_{\color{green!60!black}\text{+}0.6}$& \textbf{24.3}\tiny$_{\color{green!60!black}\text{+}0.7}$& 40.9\tiny$_{\color{red}\text{-}2.3}$& \textbf{16.2}\tiny$_{\color{gray}0.0}$& 20.6\tiny$_{\color{red}\text{-}1.5}$& \underline{52.5}\tiny$_{\color{green!60!black}\text{+}2.4}$ \\
      \rowcolor{Size14BColor} 6 & \underline{87.2}\tiny$_{\color{green!60!black}\text{+}3.7}$& \underline{80.5}\tiny$_{\color{green!60!black}\text{+}3.7}$& 84.4\tiny$_{\color{green!60!black}\text{+}8.5}$& 71.7\tiny$_{\color{green!60!black}\text{+}7.7}$& \underline{53.3}\tiny$_{\color{gray}0.0}$& 23.0\tiny$_{\color{red}\text{-}0.6}$& \textbf{43.5}\tiny$_{\color{green!60!black}\text{+}0.3}$& \textbf{16.2}\tiny$_{\color{gray}0.0}$& \textbf{22.9}\tiny$_{\color{green!60!black}\text{+}0.8}$& 51.6\tiny$_{\color{green!60!black}\text{+}1.5}$ \\
      \rowcolor{HeaderColor}\multicolumn{11}{c}{\rule{0pt}{1.8ex}\textcolor{white}{\textbf{Ucoder-32B}}\rule[-0.6ex]{0pt}{0pt}} \\
      \rowcolor{BaselineColor} 0 & 86.0& 78.0& 86.2& 72.2& 54.8& \textbf{28.4}& 44.6& \textbf{18.9}& \underline{22.1}& 53.0 \\
      \rowcolor{Size32BColor} 1 & 87.8\tiny$_{\color{green!60!black}\text{+}1.8}$& \underline{82.3}\tiny$_{\color{green!60!black}\text{+}4.3}$& \textbf{89.9}\tiny$_{\color{green!60!black}\text{+}3.7}$& \textbf{75.9}\tiny$_{\color{green!60!black}\text{+}3.7}$& \underline{55.4}\tiny$_{\color{green!60!black}\text{+}0.6}$& 23.6\tiny$_{\color{red}\text{-}4.8}$& 44.5\tiny$_{\color{red}\text{-}0.1}$& \underline{17.6}\tiny$_{\color{red}\text{-}1.3}$& 17.6\tiny$_{\color{red}\text{-}4.5}$& \underline{54.3}\tiny$_{\color{green!60!black}\text{+}1.2}$ \\
      \rowcolor{Size32BColor} 2 & 86.6\tiny$_{\color{green!60!black}\text{+}0.6}$& 81.1\tiny$_{\color{green!60!black}\text{+}3.1}$& 88.4\tiny$_{\color{green!60!black}\text{+}2.2}$& 74.6\tiny$_{\color{green!60!black}\text{+}2.4}$& \textbf{56.2}\tiny$_{\color{green!60!black}\text{+}1.4}$& 23.0\tiny$_{\color{red}\text{-}5.4}$& \underline{44.8}\tiny$_{\color{green!60!black}\text{+}0.2}$& \textbf{18.9}\tiny$_{\color{gray}0.0}$& 16.0\tiny$_{\color{red}\text{-}6.1}$& 52.4\tiny$_{\color{red}\text{-}0.6}$ \\
      \rowcolor{Size32BColor} 3 & 87.8\tiny$_{\color{green!60!black}\text{+}1.8}$& 81.1\tiny$_{\color{green!60!black}\text{+}3.1}$& 88.9\tiny$_{\color{green!60!black}\text{+}2.7}$& 74.6\tiny$_{\color{green!60!black}\text{+}2.4}$& 54.3\tiny$_{\color{red}\text{-}0.5}$& 22.3\tiny$_{\color{red}\text{-}6.1}$& 43.8\tiny$_{\color{red}\text{-}0.8}$& 16.9\tiny$_{\color{red}\text{-}2.0}$& 19.8\tiny$_{\color{red}\text{-}2.3}$& \textbf{54.7}\tiny$_{\color{green!60!black}\text{+}1.7}$ \\
      \rowcolor{BestIterColor} \textbf{4} & \textbf{89.0}\tiny$_{\color{green!60!black}\text{+}3.0}$& \textbf{82.9}\tiny$_{\color{green!60!black}\text{+}4.9}$& \underline{89.7}\tiny$_{\color{green!60!black}\text{+}3.5}$& \underline{75.7}\tiny$_{\color{green!60!black}\text{+}3.5}$& \underline{55.4}\tiny$_{\color{green!60!black}\text{+}0.6}$& \underline{27.7}\tiny$_{\color{red}\text{-}0.7}$& \textbf{45.7}\tiny$_{\color{green!60!black}\text{+}1.1}$& \underline{17.6}\tiny$_{\color{red}\text{-}1.3}$& 21.4\tiny$_{\color{red}\text{-}0.7}$& 53.4\tiny$_{\color{green!60!black}\text{+}0.3}$ \\
      \rowcolor{Size32BColor} 5 & \underline{88.4}\tiny$_{\color{green!60!black}\text{+}2.4}$& 81.7\tiny$_{\color{green!60!black}\text{+}3.7}$& 87.8\tiny$_{\color{green!60!black}\text{+}1.6}$& 73.3\tiny$_{\color{green!60!black}\text{+}1.1}$& 54.5\tiny$_{\color{red}\text{-}0.3}$& 24.3\tiny$_{\color{red}\text{-}4.1}$& 43.8\tiny$_{\color{red}\text{-}0.8}$& \textbf{18.9}\tiny$_{\color{gray}0.0}$& \textbf{22.9}\tiny$_{\color{green!60!black}\text{+}0.8}$& 53.3\tiny$_{\color{green!60!black}\text{+}0.2}$ \\
      \rowcolor{Size32BColor} 6 & \underline{88.4}\tiny$_{\color{green!60!black}\text{+}2.4}$& \underline{82.3}\tiny$_{\color{green!60!black}\text{+}4.3}$& 89.2\tiny$_{\color{green!60!black}\text{+}3.0}$& 74.1\tiny$_{\color{green!60!black}\text{+}1.9}$& 54.0\tiny$_{\color{red}\text{-}0.8}$& 23.6\tiny$_{\color{red}\text{-}4.8}$& 44.1\tiny$_{\color{red}\text{-}0.5}$& 16.9\tiny$_{\color{red}\text{-}2.0}$& \textbf{22.9}\tiny$_{\color{green!60!black}\text{+}0.8}$& 53.8\tiny$_{\color{green!60!black}\text{+}0.7}$ \\
      \bottomrule
    \end{tabular}%
  }%
  \caption{Performance (Pass@1) across iterative SFT rounds at different model scales using Qwen2.5-Coder as the base model. Orange-highlighted rows show Iter 0 (initial model trained on seed data), while subsequent iterations use self-generated synthetic data. Blue-highlighted rows indicate the best-performing iteration for each scale. Bold denotes best performance, underline denotes second-best performance for each metric within each scale, and subscripts show differences from Iter 0 (${\color{green!60!black}green}$ for improvement, ${\color{red}red}$ for decline).}
  \label{tab:iterative_sft}
  \vspace{-15pt}
\end{table*}

We select the valid candidates using three criteria:

\noindent(1) Reliability Filtering.
Candidates with low execution success are removed (threshold $\rho=0.8$): $R'=\{r\in R:e(r)\ge\rho\}$.

\noindent(2) Consensus Selection.
We select the largest non-trivial cluster: $C^*=\arg\max_{C\in\mathcal{C}',\,|C|\ge\tau}|C|$

\noindent(3) Intra-Cluster Selection.
Within $C^*$, we choose $r^*=\arg\max_{r\in C^*}\langle e(r),-f(r)\rangle$.

\subsubsection{Theoretical Guarantee}

\begin{theorem}[Consensus Convergence]
\label{thm:consensus}
Let $R=\{r_1,\dots,r_n\}$ be $n$ candidates sampled independently from a model, and let T denote a set of unit tests. 
Assume that at least $k$ candidates in $R$ are functionally correct with probability at least $1-\delta$, and that any pair of incorrect implementations produces identical outputs on a single test with probability at most $p<1$.

If the test set size satisfies
$$
|T| \ge \frac{\log(n/k)}{-\log p},
$$
then the largest consensus cluster $C_{\max}$ contains only correct implementations with probability at least
$$
P(C_{\max}\ \text{is correct}) \ge 1-\delta-n^2p^{|\mathcal{T}|}.
$$
\end{theorem}

\subsection{Iterative Self-Training}

We formalize the iterative self-training procedure and explain why it yields consistent improvement.

\begin{definition}[Iterative Update]
At iteration $t$, we construct training set $\mathcal{D}_t = \{(q_i, r_i^*)\}$, where each $r_i^*$ is selected via consensus from $n$ candidates sampled from $\mathcal{M}_t$, and update:
\begin{MiddleEquation}
\begin{equation}
\theta_{t+1} = \arg\max_\theta \sum_{(q, r^*) \in \mathcal{D}_t} \log p_\theta(r^* \mid q).
\end{equation}
\end{MiddleEquation}
\end{definition}

\paragraph{Why Self-Training Improves Performance?}
Iterative self-training is effective because consensus selection acts as a quality filter. Let $Q(r) \in [0,1]$ denote candidate quality. For $n$ independent samples, random selection yields expected quality $\mathbb{E}_{r \sim \mathcal{M}_t}[Q(r)]$, whereas consensus selection favors correct implementations that cluster by execution behavior, achieving (for some $\Delta > 0$):
\begin{MiddleEquation}\begin{equation}
\mathbb{E}[Q(r^*)] = \mathbb{E}_{r \sim \mathcal{M}_t}[Q(r)] + \Delta,
\end{equation}\end{MiddleEquation}

Optimizing on $\mathcal{D}_t$ shifts the model toward higher-quality samples. As $\mathcal{M}_{t+1}$ increases $p_\theta(r^* \mid q)$ for above-average outputs:
\begin{MiddleEquation}
\begin{equation}
\mathbb{E}_{r \sim \mathcal{M}_{t+1}}[Q(r)] \geq \mathbb{E}_{r \sim \mathcal{M}_t}[Q(r)],
\end{equation}
\end{MiddleEquation}where it induces a positive feedback loop where improved models generate higher-quality candidates and more reliable training signals.

\section{Experiments} 
\subsection{Training and Evaluation Details}
\paragraph{Model Configuration.}
We experiment with Qwen2.5-Coder\cite{qwen25coder} models at 7B, 14B, and 32B scales, starting from base checkpoints without prior instruction tuning and applying identical self-bootstrapping procedures across all scales for fair comparison.

\paragraph{Training Hyperparameters.}
We employ consistent training settings across experiments. Models are fine-tuned for 3 epochs per iteration using AdamW with a learning rate of 5e-6 and a cosine decay schedule. We use a batch size of 128 with gradient accumulation to fit memory constraints.

\paragraph{Evaluation Benchmarks.}
We evaluate on six benchmarks: \textbf{HumanEval}~\cite{humaneval} and \textbf{MBPP/MBPP+}~\cite{mbpp} assess classic Python programming; \textbf{LiveCodeBench}~\cite{LiveCodeBench} provides contamination-free competitive programming problems; \textbf{BigCodeBench (BCB)}~\cite{bigcodebench} evaluates function completion with broader context and API usage (both Complete and Instruct variants); and \textbf{FullStackBench}~\cite{fullstack} covers diverse real-world scenarios. We report Pass@1 accuracy with execution-based validation; solutions must pass all test cases.



\begin{figure}[t]
    \centering
    \includegraphics[width=\linewidth]{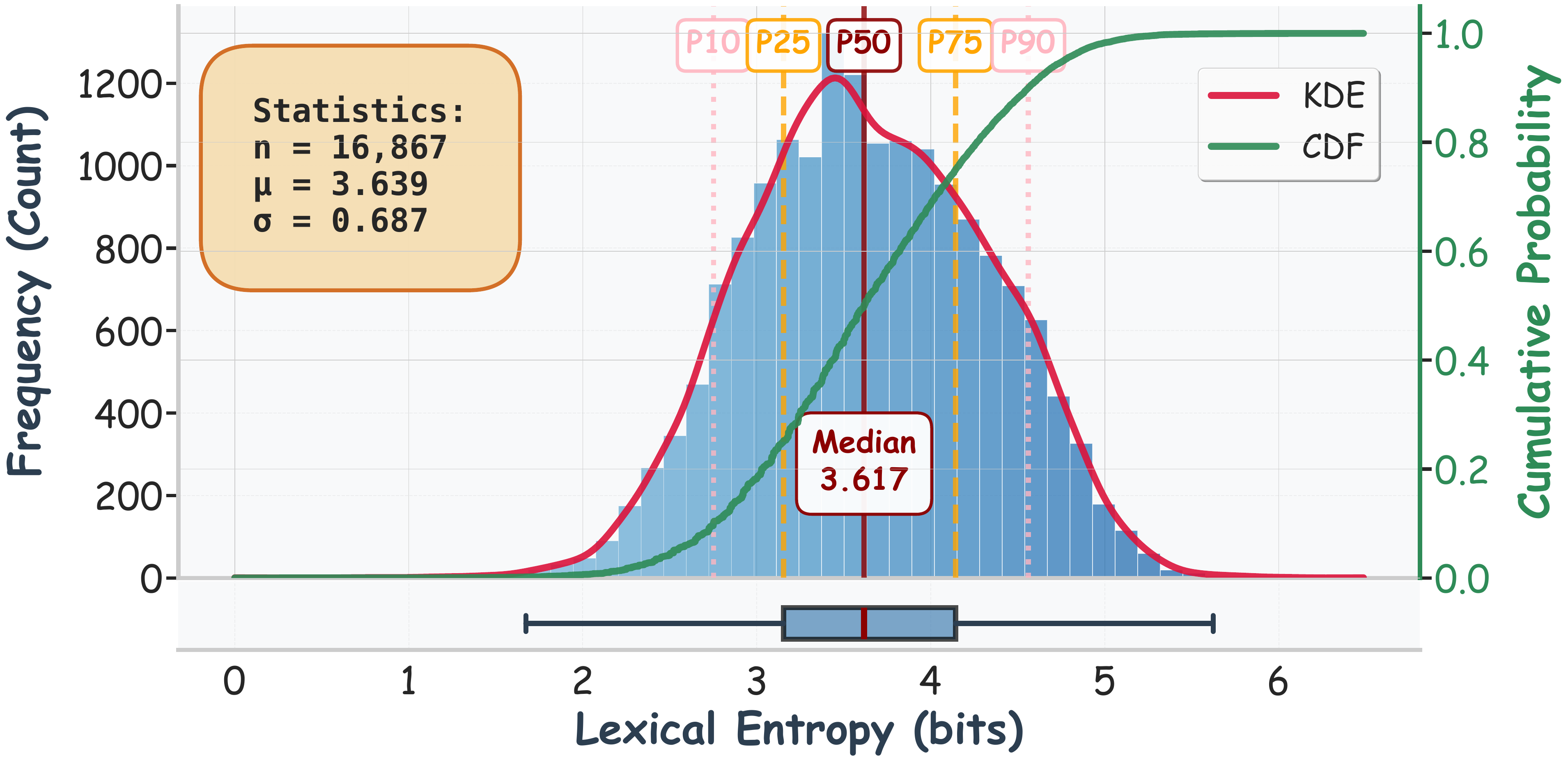}
    \caption{Lexical entropy distribution of 16,867 generated problems. Histogram with KDE shows per-problem entropy; CDF (green) and boxplot show cumulative coverage.}
    \label{fig:lexical}
    \vspace{-20pt}
\end{figure}

\begin{figure}[t]
    \centering
    \includegraphics[width=\linewidth]{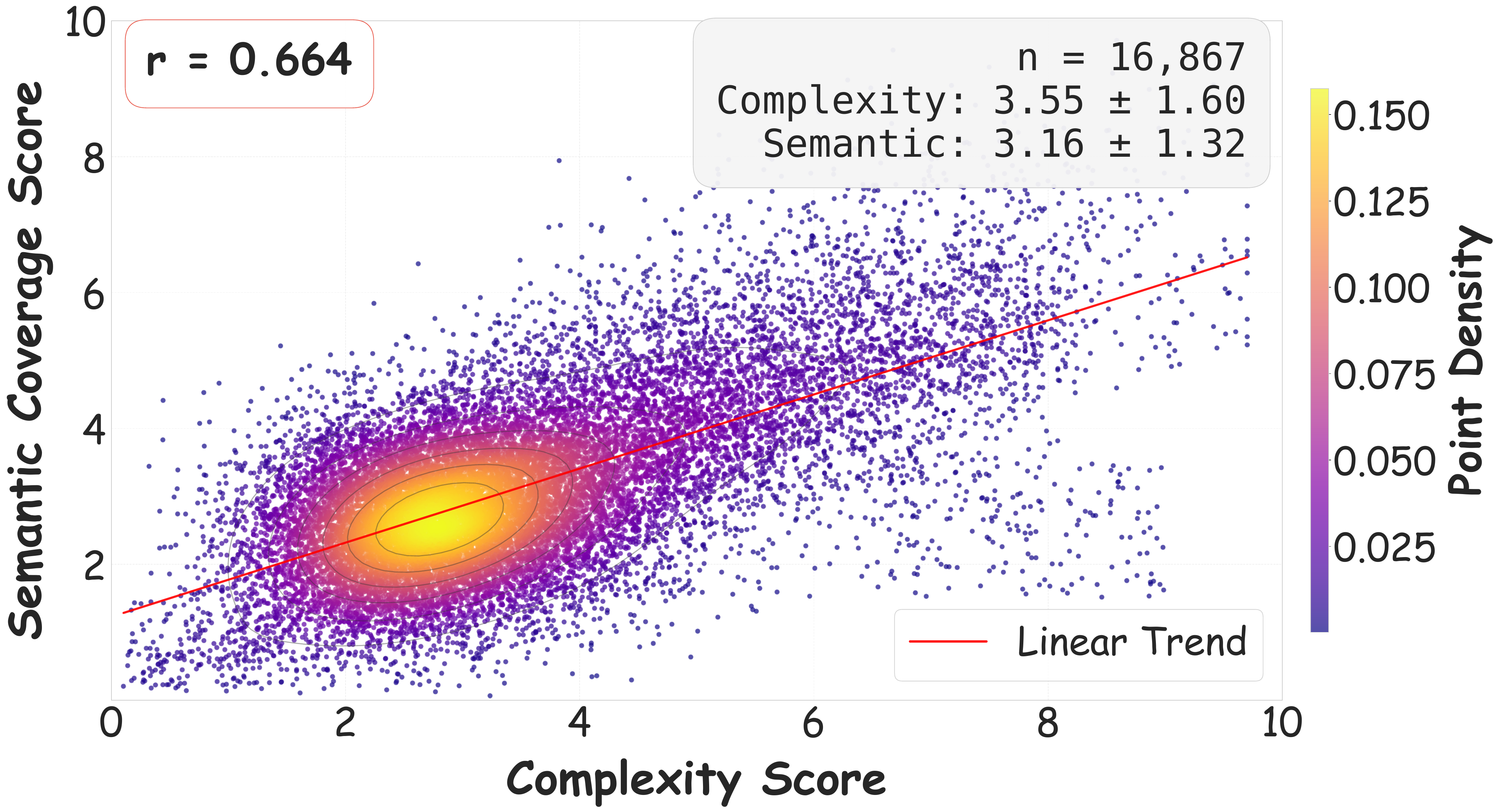}
    \caption{Complexity versus semantic coverage distribution. Color encodes density; red line shows linear trend ($r = 0.664$).}
    \vspace{-10pt}
    \label{fig:complexity}
\end{figure}

\begin{figure}[t]
    \centering
    \includegraphics[width=\linewidth]{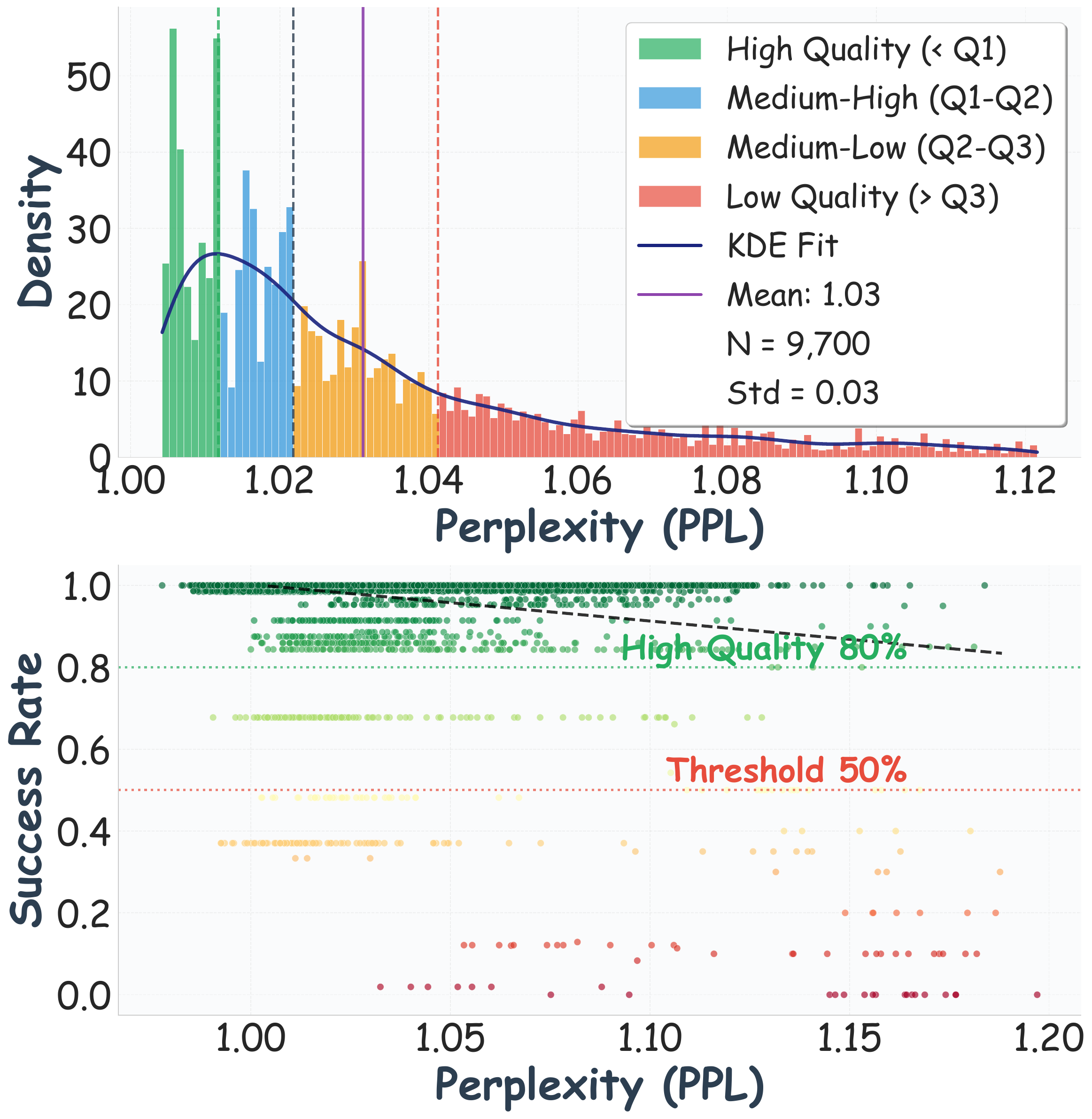}
    \caption{Quality distribution characterization. 
    Top: perplexity distribution across 9,700 samples with quartile stratification and KDE overlay (truncated at 1.12). Bottom: perplexity versus execution success rate, showing high-quality samples (80\%+ success) concentrated below perplexity 1.05.}
    \label{fig:quality_distribution}
    \vspace{-15pt}
\end{figure}

\subsection{Main Results}
\label{subsec:results}


\autoref{tab:qwen25_base_instruct_comparison} demonstrates that unsupervised self-bootstrapping achieves performance comparable to supervised instruction tuning across diverse code generation benchmarks. Compared with other instruction-tuned models of similar scale, \sftmodel{} exhibits competitive or superior performance, consistently matching or exceeding Qwen2.5-Coder-Instruct on challenging benchmarks including MBPP+, BigCodeBench-Complete, and FullStackBench at all model scales (7B, 14B, 32B). Although certain Instruct models retain advantages on HumanEval, our approach progressively narrows this gap as model scale increases. These results indicate that execution-driven self-training can effectively elicit the latent instruction-following capabilities embedded in pre-trained code models, achieving supervised-level performance without requiring any human annotations or curated instruction data.

\subsection{Effects of Iterative Self-Improvement}
\label{subsec:iterative}

\autoref{tab:iterative_sft} reports performance across six self-bootstrapping iterations at three model scales, demonstrating both framework effectiveness and scale-dependent dynamics.

\paragraph{Framework Effectiveness.}
Iterative self-bootstrapping consistently improves performance without external supervision. Across benchmarks, all model scales show substantial gains over seed-trained baselines (Iter 0), with improvements of +6.1 to +13.2 points at 7B, +4.3 to +10.6 at 14B, and +3.0 to +4.9 at 32B. Gains are most pronounced on benchmarks requiring diverse programming skills, such as MBPP, FullStackBench, and LiveCodeBench, rather than narrowly scoped tasks like HumanEval. This supports our hypothesis that self-generated problem diversity combined with execution-driven consensus expands capability coverage beyond seed data.

\paragraph{Inverse Scaling of Improvement.}
Performance gains exhibit an inverse scaling trend, with smaller models benefiting disproportionately. We attribute this effect to \emph{latent capability gaps}, where pre-trained knowledge is only partially accessible through standard prompting. Consensus-based selection mitigates this gap by reinforcing correct patterns that smaller models generate inconsistently. Notably, the self-improved 7B model reaches 85.2\% on MBPP, approaching the 32B baseline (86.2\%), highlighting self-bootstrapping as a compute-efficient alternative to model scaling.

\paragraph{Convergence Characteristics.}
The optimal number of iterations decreases with model scale—six for 7B, five for 14B, and four for 32B. Beyond these points, performance exhibits mild oscillation rather than degradation, reflecting a trade-off between specialization on synthetic data and generalization to held-out distributions. These observations motivate early stopping based on validation performance.

\definecolor{HeaderColor}{HTML}{9CA3AF}        
\definecolor{Size7BColor}{HTML}{F3E5F5}        
\definecolor{Size14BColor}{HTML}{E8F5E8}       
\definecolor{Size32BColor}{HTML}{F9FBE7}       
\definecolor{OursColor}{HTML}{E3F2FD}          


\begin{table*}[t]
  \centering
  \small
  \renewcommand{\arraystretch}{1.3}
  \setlength{\tabcolsep}{4pt}
  \resizebox{0.9\textwidth}{!}{%
    \begin{tabular}{l|CC|CC|CC|CC|c|c}
        \toprule
      \multirow{2}{*}{\textbf{Method}} & 
      \multicolumn{2}{c|}{\textbf{HumanEval}} & 
      \multicolumn{2}{c|}{\textbf{MBPP}} & 
      \multicolumn{2}{c|}{{\textbf{BCB-Complete}}} & 
      \multicolumn{2}{c|}{{\textbf{BCB-Instruct}}} & 
      \textbf{LiveCode-} & 
      \textbf{FullStack-} \\
      & HE & HE+ & MBPP & MBPP+ & Full & Hard & Full & Hard & \textbf{Bench} & \textbf{Bench} \\
      \rowcolor{HeaderColor}\multicolumn{11}{c}{\rule{0pt}{2.5ex}\textcolor{white}{\textbf{7B Models}}\rule[-1ex]{0pt}{1ex}} \\
      \rowcolor{OursColor} \textbf{\sftmodel{}} & \textbf{77.4} & 67.1 & \textbf{72.0} & \textbf{63.0} & \underline{44.4} & \underline{15.5} & 34.6 & \textbf{14.9} & \textbf{13.0} & \textbf{40.25} \\
      \rowcolor{Size7BColor} Random & 73.8 & 64.6 & 65.6 & 56.6 & \underline{45.5} & 14.9 & \textbf{37.5} & \underline{13.5} & 10.7 & 33.55 \\
      \rowcolor{Size7BColor} Cluster & 73.8 & 66.5 & \underline{68.8} & 58.5 & 41.3 & 12.2 & 36.5 & 12.2 & 10.7 & 38.35 \\
      \rowcolor{Size7BColor} Low PPL & \textbf{77.4} & \textbf{70.7} & 70.1 & \underline{59.3} & \textbf{45.6} & \textbf{17.6} & \underline{37.3} & 12.2 & \underline{12.2} & 37.11 \\
      \rowcolor{Size7BColor} Success Rate & \underline{75.0} & 67.7 & 66.9 & 57.1 & 41.4 & 11.5 & 33.0 & 12.2 & \underline{12.2} & \underline{40.01} \\
      \rowcolor{HeaderColor}\multicolumn{11}{c}{\rule{0pt}{2.5ex}\textcolor{white}{\textbf{14B Models}}\rule[-1ex]{0pt}{1ex}} \\
      \rowcolor{OursColor} \textbf{\sftmodel{}} & \textbf{83.5} & \underline{76.8} & \textbf{75.9} & \textbf{64.0} & \underline{53.3} & \textbf{23.6} & \underline{43.2} & 16.2 & \textbf{22.1} & \textbf{50.09} \\
      \rowcolor{Size14BColor} Random & 81.7 & 76.2 & 71.7 & 60.1 & \textbf{53.8} & \underline{20.9} & \textbf{43.3} & \underline{17.6} & 16.8 & 47.95 \\
      \rowcolor{Size14BColor} Cluster & 82.3 & \underline{76.8} & 73.0 & \textbf{64.0} & 50.4 & 18.2 & 42.4 & 15.5 & 17.6 & 48.55 \\
      \rowcolor{Size14BColor} Low PPL & \underline{82.9} & \textbf{77.4} & 73.3 & \underline{63.5} & 50.7 & 20.3 & 42.2 & \textbf{23.0} & \underline{19.1} & \underline{49.32} \\
      \rowcolor{Size14BColor} Success Rate & 82.3 & 75.6 & \underline{74.1} & 63.2 & 50.6 & \underline{20.9} & 41.3 & 14.9 & 17.6 & 47.36 \\
      \rowcolor{HeaderColor}\multicolumn{11}{c}{\rule{0pt}{2.5ex}\textcolor{white}{\textbf{32B Models}}\rule[-1ex]{0pt}{1ex}} \\
      \rowcolor{OursColor} \textbf{\sftmodel{}} & \textbf{86.0} & \underline{78.0} & \textbf{86.2} & \textbf{72.2} & \textbf{54.8} & \textbf{28.4} & 44.6 & 18.9 & \textbf{22.1} & \textbf{53.05} \\
      \rowcolor{Size32BColor} Random & \underline{84.8} & \textbf{79.3} & 80.7 & 69.6 & 53.7 & 25.0 & \textbf{46.6} & \underline{21.6} & 15.3 & 39.18 \\
      \rowcolor{Size32BColor} Cluster & 83.5 & 75.6 & \underline{81.5} & \underline{71.2} & 53.9 & 23.0 & 45.1 & 16.9 & 13.0 & 49.91 \\
      \rowcolor{Size32BColor} Low PPL & 84.1 & 76.2 & 79.9 & 67.5 & 53.5 & \underline{27.7} & \underline{45.4} & \textbf{23.0} & 18.3 & \underline{50.09} \\
      \rowcolor{Size32BColor} Success Rate & 82.3 & 75.6 & 81.2 & 69.6 & \underline{54.1} & \underline{27.7} & 44.2 & 17.6 & \underline{21.4} & 50.50 \\
       \bottomrule
    \end{tabular}%
  }%
  \caption{Ablation study comparing data selection strategies across model scales: \textbf{\sftmodel{}} (execution-driven consensus), \textbf{Random} (random sampling from successful solutions), Cluster (clustering-based), \textbf{Low PPL} (lowest perplexity), and \textbf{Success Rate} (weighted by execution success). All metrics show Pass@1 execution rates (\%). \textbf{Bold/Underline} indicate best/second-best performance per size category.}
  \label{tab:selection_strategy_comparison}
  \vspace{-15pt}
\end{table*}

\section{Analysis}
\subsection{Diversity of Self-Generated Problems}
\label{sec:analysis}

\paragraph{Lexical Diversity.}
We quantify lexical diversity using Shannon entropy $H = -\sum_i p_i \log_2 p_i$, where $p_i$ denotes token probability. Figure~\ref{fig:lexical} shows a near-Gaussian entropy distribution (mean $\mu = 3.64$ bits, $\sigma = 0.69$, median 3.62), indicating natural variation rather than templated construction. The smooth CDF and moderate interquartile range further suggest balanced coverage from concise to elaborate specifications.

\paragraph{Semantic Coverage.}
As shown in Figure~\ref{fig:wordcloud} (Appendix), the generated problems contain 229 domain-specific terms across seven categories, with Data Structures (18.3\%), Algorithms (14.8\%), and String Processing (11.4\%) being most prominent. No category exceeds 20\%, demonstrating broad semantic coverage, while concrete algorithmic terms (e.g., \textit{dijkstra}, \textit{greedy}, \textit{traversal}) indicate non-generic, verifiable challenges.

\paragraph{Complexity Distribution.}
We assess problem difficulty and conceptual breadth using a \textit{Complexity Score} (0–10, aggregating parameter count, description length, algorithmic keywords, and constraints) and a \textit{Semantic Coverage Score} (weighted keyword matches across seven categories). Figure~\ref{fig:complexity} shows a moderate positive correlation ($r = 0.664$), indicating that more complex problems tend to integrate multiple concepts. Problems span the full score ranges (complexity: $3.55 \pm 1.60$, semantic: $3.16 \pm 1.32$) with a continuous density profile, suggesting a natural difficulty continuum suitable for curriculum learning.



\subsection{Execution-Driven Consensus Effectiveness}
\label{sec:consensus_analysis}


\paragraph{Solution Space Diversity.}
We first examine whether dense sampling with $n=128$ candidates yields a sufficiently diverse solution pool. As shown in Figure~\ref{fig:solution_diversity_app}, the abstract syntax tree (AST) node distribution spans 15 syntactic constructs across 2.6M generated samples (212M total nodes), indicating diverse implementation structures beyond primitive elements. The joint distribution of cyclomatic complexity and code length further shows broad dispersion (mean complexity $2.7 \pm 2.3$, mean length $22.4 \pm 10.5$ lines), confirming substantial heterogeneity in the solution space.

\paragraph{Quality Distribution Characterization.}
Given this diversity, we examine whether solution quality exhibits sufficient separation for reliable selection. Figure~\ref{fig:quality_distribution} shows the perplexity distribution over 9,700 sampled candidates and its relationship with execution success. The distribution is right-skewed (mean 1.03, std 0.03) with clear stratification: high-quality samples concentrate at low perplexity values around 1.01, while lower-quality samples progressively shift toward higher perplexity, extending beyond 1.10. Consistently, solutions with execution success rates above 80\% cluster predominantly below perplexity 1.05, with rapid performance degradation beyond this range. This sharp transition indicates that high-quality solutions form a distinct and identifiable subset within the candidate pool.
\begin{figure}[h]
    \centering
    \includegraphics[width=\linewidth]{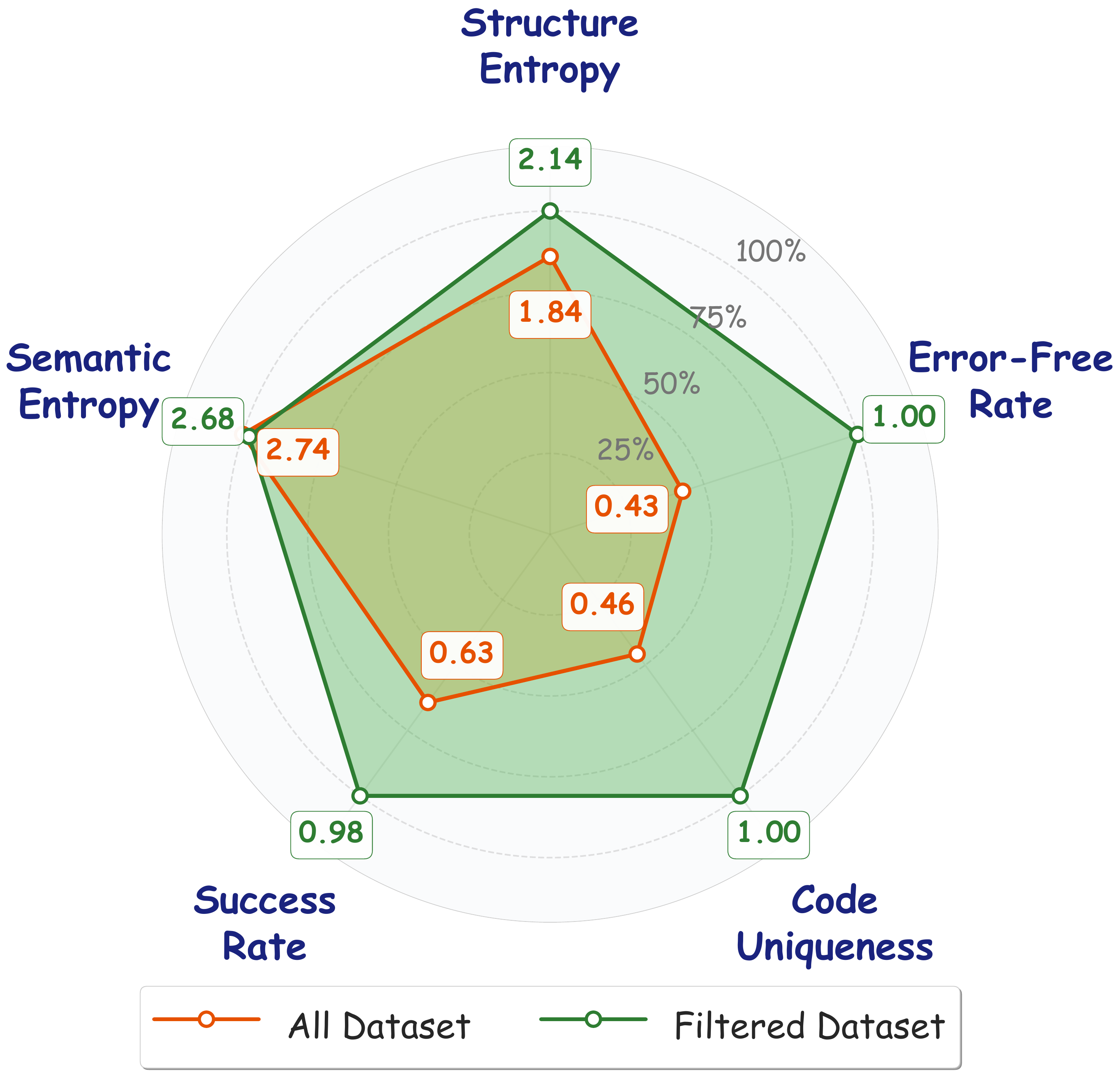}
    \caption{Filtered dataset (green) improves quality over full dataset (gray) while maintaining diversity.}
    \vspace{-20pt}
    \label{fig:filtering_effectiveness}
\end{figure}

\paragraph{Quality Improvement.}
Consensus-based filtering yields substantial improvements across all quality metrics. In \autoref{fig:filtering_effectiveness}, the filtered dataset consistently outperforms the full dataset in success rate, error-free rate, and code uniqueness. Besides, these gains are achieved without sacrificing diversity, where structural and semantic entropy remain comparable between the two datasets. This demonstrates that execution-driven consensus selection effectively resolves the quality–diversity trade-off.



\subsection{Ablation Study}
\label{subsec:ablation}

To isolate the effect of data selection, we compare our consensus-based approach with four alternatives: \textbf{Random} uniform sampling, \textbf{Cluster} selection from dominant output-hash clusters, \textbf{Low PPL} selection based on minimal perplexity, and \textbf{Success Rate} filtering with a 0.5 execution threshold. \autoref{tab:selection_strategy_comparison} shows our approach consistently outperforms all baselines across three model scales, achieving the best results on most benchmarks. The performance gap widens with scale, on FullStackBench, the margin over Random increases from 6.7 points at 7B to 13.9 points at 32B, suggesting larger models produce more diverse candidate pools where principled selection matters most. While individual baselines excel on specific benchmarks, none matches the robustness of consensus-based selection. Notably, Success Rate filtering fails to consistently outperform Random, indicating binary pass/fail criteria inadequately capture quality differences, whereas consensus-based selection leverages behavioral consistency across test inputs as a richer quality signal.

\section{Related Work}
\paragraph{Unsupervised Learning for Code Post-training.} 
Unsupervised learning has become increasingly prominent in code generation through pre-training on vast unlabeled code repositories, building on early findings that source code exhibits statistical regularities similar to natural language~\cite{code_llama,deepseek_coder,starcoder,starcoder2,vgamegym,livereporeflection}. Recent work on unsupervised code post-training has focused on leveraging unlabeled code snippets and use LLM to generate synthetic question-answering data. Magicoder~\cite{magicoder} uses open-source code examples to teach LLMs how to create varied coding instructions and training data. Besides, WizardCoder~\cite{wizardcoder} progressively evolves simple coding instructions into complex ones for training. Further, WaveCoder~\cite{codearena} and CodeArena~\cite{codearena} generate more diverse and high-quality instruction data from the open source code dataset. Besides, there are some works~\cite{unsupervised_code_changes,unsupervised_code_translation} that adopt supervised learning for code translation and code change tasks.

\paragraph{Code Instruction Tuning.}
Instruction tuning~\cite{opencoder,qwen25coder, yang2025codefoundationmodelsagents} has emerged as an effective approach for improving LLMs by fine-tuning them on instruction-based datasets, enabling better generalization and more accurate instruction-following capabilities. To improve the capabilities of code LLM, researchers have enhanced multiple code tasks and benchmarks, such as code generation of multiple domains (e.g., BigCodeBench~\cite{bigcodebench} and FullStackBench~\cite{fullstack}), multilingual code generation (e.g., MultiPl-E~\cite{multiple} and McEval~\cite{mceval}), and competitive programming problems (e.g., LiveCodeBench~\cite{LiveCodeBench}).

\section{Conclusion}
In this work, we introduce an execution-driven self-bootstrapping framework that removes the need for human-annotated instruction data in code generation. UCoder exploits latent programming knowledge in pre-trained models and uses execution feedback as a scalable, deterministic supervision signal for autonomous improvement. Guided by execution-driven consensus clustering, iterative self-training identifies correct implementations via behavioral consistency and constructs high-quality training data. Our results demonstrate that code models can achieve performance competitive with supervised baselines through fully autonomous learning, highlighting a scalable and cost-effective path for advancing code intelligence.

\clearpage
\section*{Limitations}
Despite the promising results, our work has several limitations that warrant future investigation. First, the effectiveness of our consensus-based selection mechanism relies on the availability of executable test cases, which may not always be feasible for certain programming tasks or domains where formal specifications are difficult to construct. Second, while our method demonstrates strong performance on standard benchmarks, the computational cost of generating and evaluating 128 candidate solutions per problem remains substantial, potentially limiting its applicability in resource-constrained scenarios. Third, our approach primarily focuses on functional correctness through execution-based validation, which may not capture other important code quality attributes such as maintainability, documentation quality, or adherence to specific coding standards beyond those explicitly testable. Fourth, the iterative self-training process exhibits diminishing returns and potential overfitting to synthetic data distributions after a certain number of iterations, requiring careful validation-based early stopping. Finally, our analysis is primarily conducted on Python programming tasks, and the generalizability of our findings to other programming languages with different execution characteristics and paradigms remains to be thoroughly validated. These limitations highlight important directions for future work in unsupervised code generation.

\section*{Ethics Statement}
This work on unsupervised code generation acknowledges several ethical considerations. Automated code generation systems can be misused to produce malicious code or security vulnerabilities, requiring appropriate safeguards, including content filtering and usage monitoring. All experiments used publicly available benchmarks and open-source models, ensuring transparency without collecting proprietary data. We advocate for responsible development with clear documentation, transparent methodologies, and adherence to intellectual property rights.

\section*{Acknowledgment}
This work was supported by State Key Laboratory of Complex \& Critical Software Environment (SKLCCSE) of Beihang University and supported by the Fundamental Research Funds for the Central Universities. This work was supported in part by the National Natural Science Foundation of China (Grant Nos. 62276017, 62406033, U1636211, 61672081), and the State Key Laboratory of Complex \& Critical Software Environment (Grant No. SKLCCSE-2024ZX-18).

\bibliography{custom}
\bibliographystyle{acl_natbib}

\clearpage

\appendix
\onecolumn
\newpage
\section{Additional Analysis Figures}

\begin{figure}[h]
    \centering
    \includegraphics[width=0.8\linewidth]{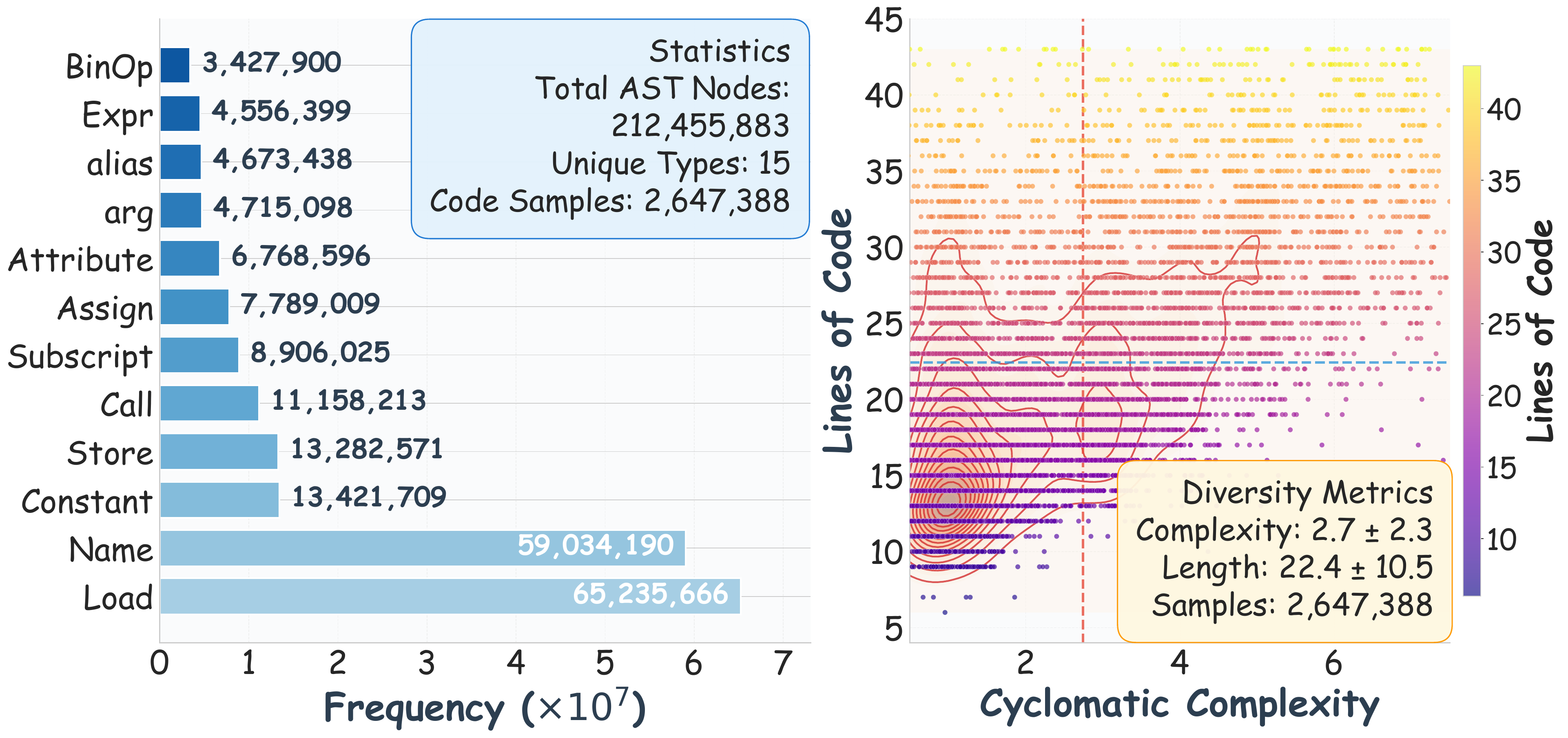}
    \caption{Solution space diversity analysis. Left: AST node type distribution across 2.6M samples totaling 212M nodes, spanning 15 distinct syntactic constructs. Right: Cyclomatic complexity versus code length with density contours, showing broad dispersion (complexity: $2.7 \pm 2.3$, length: $22.4 \pm 10.5$ lines).}
    \label{fig:solution_diversity_app}
\end{figure}

\begin{figure}[!b]
    \centering
    \includegraphics[width=0.8\linewidth]{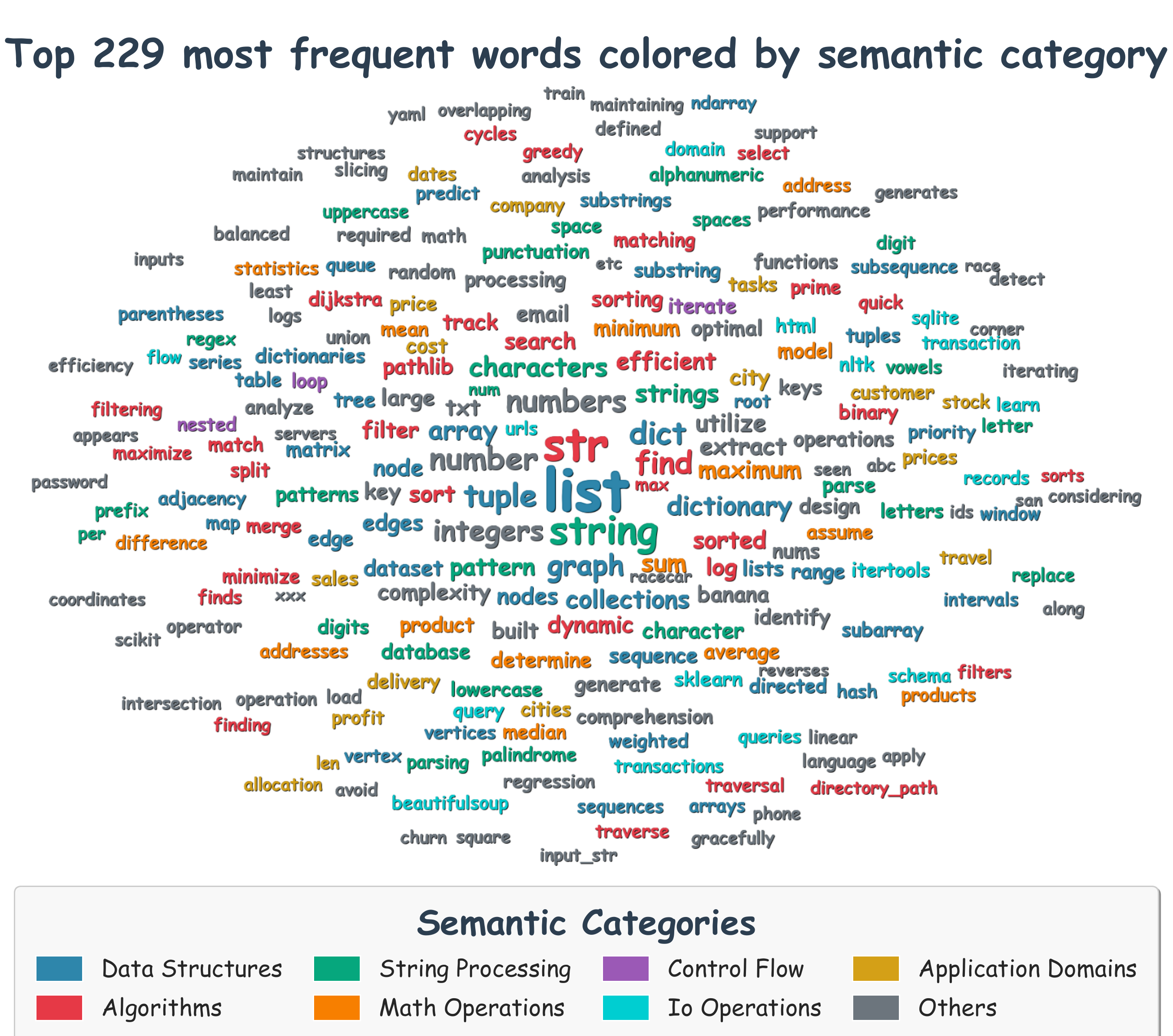}
    \caption{Semantic distribution of 229 frequent terms in generated problems. Word size indicates frequency; colors denote categories.}
    \label{fig:wordcloud}
\end{figure}

\end{CJK*}
\end{document}